\documentclass{article}

    \PassOptionsToPackage{numbers, compress}{natbib}

    \usepackage[final]{neurips_2024}

\usepackage[utf8]{inputenc} 
\usepackage[T1]{fontenc}    
\usepackage{hyperref}       
\usepackage{url}            
\usepackage{booktabs}       
\usepackage{amsfonts}       
\usepackage{nicefrac}       
\usepackage{microtype}      
\usepackage{xcolor}         

\usepackage{amsmath}
\usepackage{algorithm}
\usepackage{algorithmic}
\usepackage{graphicx}
\usepackage{booktabs}
\usepackage{tcolorbox}
\usepackage{multirow}
\usepackage{svg}

\title{Efficient Alignment of Large Language Models via Data Sampling}

\author{%
  Amrit Khera$^{1\thanks{Work done during internship at Nutanix}}$, Rajat Ghosh$^2$, Debojyoti Dutta$^2$ \\
  $^1$Georgia Institute of Technology, $^2$Nutanix \\
  \texttt{akhera30@gatech.edu, \{rajat.ghosh, debojyoti.dutta\}@nutanix.com} \\
}

\begin{document}

\maketitle

\begin{abstract}
LLM alignment ensures that large language models behave safely and effectively by aligning their outputs with human values, goals, and intentions. Aligning LLMs employ huge amounts of data, computation, and time. Moreover, curating data with human feedback is expensive and takes time. Recent research depicts the benefit of data engineering in the fine-tuning and pre-training paradigms to bring down such costs. However, alignment differs from the afore-mentioned paradigms and it is unclear if data efficient alignment is feasible. In this work, we first aim to understand how the performance of LLM alignment scales with data. We find out that LLM alignment performance follows an exponential plateau pattern which tapers off post a rapid initial increase. Based on this, we identify data subsampling as a viable method to reduce resources required for alignment. Further, we propose an information theory-based methodology for efficient alignment by identifying a small high quality subset thereby reducing the computation and time required by alignment. We evaluate the proposed methodology over multiple datasets and compare the results. We find that the model aligned using our proposed methodology \textbf{outperforms} other sampling methods and performs comparable to the model aligned with the full dataset while using less than \textbf{10\% data}, leading to greater than \textbf{90\% savings} in costs, resources, and faster LLM alignment.
\end{abstract}

\section{Introduction}

Large Language Models (LLMs) trained on huge corpus of data have shown considerable performance in distilling the knowledge and acquiring a wide variety of skills and capabilities \cite{wan2023gpt}, \cite{wang2023gpt}, \cite{wei2023zero}. Usually, the model is specialized using domain adaptive training, fine-tuning or instruction tuning \cite{radford2018improving}, \cite{wei2021finetuned}, \cite{gururangan2020don}. However, this specialized model may still produce results which humans may not find desirable. Recently, aligning LLMs with human feedback has proven to significantly improve their ability to produce ethical, factual, and helpful outputs. As demonstrated by \cite{ouyang2022training} and \cite{tian2023fine}, alignment techniques have become an indispensable part of the post-training process for LLMs, ensuring that these models adhere to human values and preferences. 

Alignment datasets usually come from human feedback which is expensive to collect, curate, and standardize. Usually, alignment techniques such as Reinforcement Learning from Human Feedback (RLHF) and Direct Preference Optimization (DPO) take
pairwise preference data which is difficult to collect due to its scarcity in the real world \cite{christiano2017deep}, \cite{rafailov2024direct}, \cite{casper2023open}. Recent alignment strategies such as KTO proposed by \cite{ethayarajh2024kto} aim to overcome the problem by converting the feedback into a binary signal and achieve SOTA performance. While binary preference data is easier to collect compared to pairwise preference data, the data management cost is still substantial. 

There have been numerous studies in the LLM pre-training and fine-tuning paradigms which show that a small high-quality dataset is sufficient, contributing to lower costs \cite{zhou2024lima}, \cite{sachdeva2024train}, \cite{liu2023makes}. In this work, we explore data efficient strategies for LLM alignment to save the overall LLM alignment cost. Specifically, we aim to answer the following research questions:
\begin{itemize}
    \item How does LLM alignment scale with data? Are we using more data than necessary? (RQ1)
    \item How can we identify optimal subsamples to sufficiently align LLMs? (RQ2)
\end{itemize}

\subsection{Contributions}

The contributions of our work are twofold: first, we carry out an empirical analysis to study how the alignment performance scales with data. We find that the alignment performance follows an exponential plateau pattern indicating less data might be sufficient for alignment. Second, we validate that small high quality subsamples can sufficiently align LLMs similar to using the entire dataset saving costs and resources. Further, we propose a principled technique for selecting high-quality and diverse data samples for alignment based on information theory. The LLM aligned using the subsample performs comparable to alignment on the entire data while using less than \textbf{10\% data}, hence significantly saving resources. We show that our methodology outperforms other sampling strategies thereby pioneering efficient alignment for LLMs.

\section{Experiment Design}
\label{sec: exp des}

To explore RQs 1 and 2, we employ the SOTA Direct Alignment Algorithm (DAA) KTO \cite{ethayarajh2024kto}, based on prospect theory which models loss objectives as Human Aware Loss Functions (HALOs). We use Mistral-7B-v0.1 \cite{jiang2023mistral7b} similar to the experiments conducted by the authors of KTO. We use the Llama-3-8B-Instruct model to encode the concatenated prompt and completion with the embedding of the last token of the concatenated sequence \cite{llama3modelcard}. The embeddings are used as the input for each sampling method. We use the OpenAssitant1 dataset \cite{köpf2023openassistantconversationsdemocratizing}, the Ultrafeedback-binarized dataset \cite{tunstall2023zephyrdirectdistillationlm} processed from the Ultrafeedback dataset \cite{cui2023ultrafeedback} for aligning Zephyr \cite{tunstall2023zephyrdirectdistillationlm}, and the Anthropic Golden HH-RLHF dataset \cite{cai2024ulmaunifiedlanguagemodel} derived from the Anthropic HH-RLHF dataset \cite{ganguli2022redteaminglanguagemodels}, \cite{bai2022traininghelpfulharmlessassistant} for our evaluations. The datasets contain 28,334, 122,270, 85,074 data samples after processing them for the KTO algorithm, respectively. The datasets chosen are commonly used for alignment of LLMs and are representative of the paradigm. Moreover, they represent a wide range of data size for generalizability. 

\section{Characteristics of Alignment Performance vs Data Sample Size (RQ-1)}
\label{sec: rq1}

In this section, we aim to understand how alignment performance scales with the amount of dataset used and whether less data can provide enough signal to align the model comparable to full-sample. For this we carry out an empirical study on multiple open datasets popularly used for aligning LLMs.
\begin{figure}[!h]
    \centering
    \includegraphics[width=0.8\linewidth]{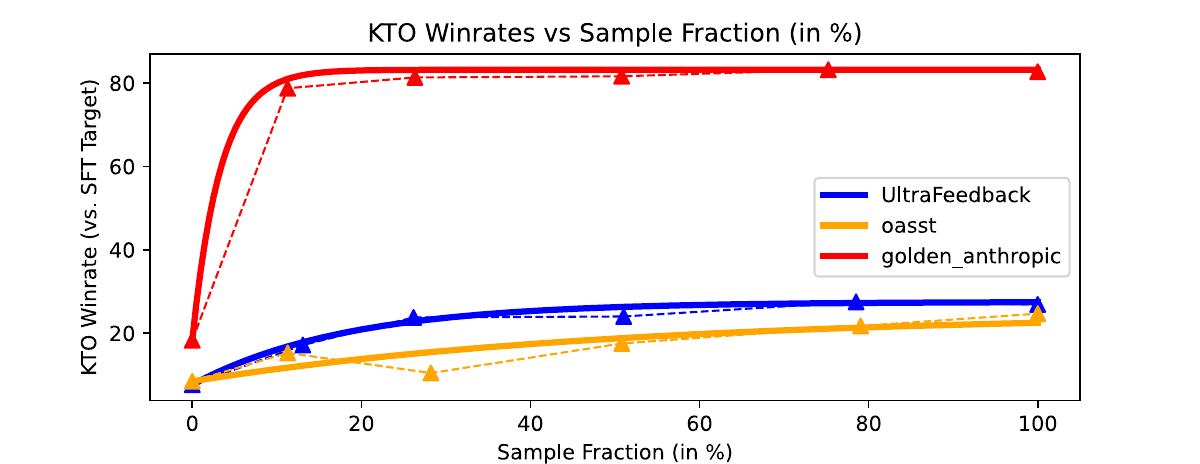}
    \caption{ Alignment data performance with Mistral-7B-v0.1. Alignment performance follows an exponential plateau pattern captured by the proposed scaling law represented by the solid lines. }
    \label{fig:laws}
\end{figure}
The results are shown in Figure \ref{fig:laws} and Table \ref{tab:data_frac}. We use GPT-4o as the Judge for the winrate computation using the same setting as the KTO study \cite{ethayarajh2024kto}. The plot and table presents the winrates of the model using intermediary checkpoints over fractions of data seen over 1 epoch of training.

From Figure \ref{fig:laws}, we can see an over-optimization trend for all three datasets as we observe an initial increase in performance which plateaus over more data. Moreover, the trend of over-optimization in KTO is different from the trend observed in contemporary work by \cite{rafailov2024scalinglawsrewardmodel}
on prior DAAs such as DPO. They observe a hump shape pattern where performance peaks around $25\%$ and then decreases as more data is seen whereas in KTO we observe a plateau in the performance after an intial increase as more data is seen. This indicates that KTO might be more robust to over-optimization than prior DAAs, however, it may be beneficial to use smaller high quality subsamples for similar alignment performance. To ensure robustness of our results in Figure \ref{fig:laws}, we ran our experiments with three seed values. Figure \ref{fig:laws} shows the mean values. As shown in Table \label{tab:data_frac} in Appendix, the 95\% confidence level for any trial did not exceed 1\%. 

Prior work in RLHF has established the scaling laws for the reward over-optimization as a function of the KL divergence between the base and aligned policies \cite{gao2023scaling}. Similarly, contemporary work has found that the same law holds for DAAs \cite{rafailov2024scalinglawsrewardmodel}. In our experiments, we observe the performance scaling as an exponential function decaying into a plateau based on the data percentage used to align the model. We treat the GPT-4o winrates over the dataset fraction as a proxy of the reward and fit curves using the empirical results over the three datasets mentioned in section \ref{sec: exp des} and the fitted curves and the data points are present in Figure \ref{fig:laws}. We observe that the curve takes the form of an exponential growth which decays into a plateau: 
\begin{equation}
 R(x) = r - (r-a)e^{-bx} \label{eq10}
\end{equation}
where, $x$ represents the percentage of the data used for alignment, $R(x)$ represents the proxy reward of GPT-4o winrates, $r$ represents the asymptote depicting the max reward attainable, $a$ represents the initial reward of the unaligned model, and $b$ represents the rate at which the reward grows which seems to be related to the complexity of the dataset. We believe both the max reward and the rate of increase depend on the complexity and characteristics of the data and aim to investigate this more thoroughly in the future. The values for the curves in Figure \ref{fig:laws} are present in Table \ref{tab: params} in the Appendix.

\section{Optimal Sample Selection for Efficient Alignment (RQ-2)}

Here, we present a novel strategy, Information Sampling for Alignment (ISA), to identify a small high quality sub-sample sufficient for alignment convergence. We first postulate that alignment data can be modeled as a Gaussian Mixture Model (GMM) Distribution. As GMM performs a soft clustering, this step can be perceived as ensuring a diverse dataset is sampled. Then we use entropy sampling from information theory to sub-sample a high quality dataset using the GMM Distribution. As alignment data is derived from human preferences, it contains either or both desired and undesired outputs for each prompt. Hence, the dataset can be considered as having two conversations: one with the desired outputs and one with the undesired ones. Hence, it makes sense to use a 2 component GMM to model the dataset. The log likelihood of a data point $x$ is given by:
\begin{equation}
 l(x) = \log[\pi\mathcal{N}(x | \mu_1, \Sigma_1) + (1-\pi)\mathcal{N}(x | \mu_2, \Sigma_2)]
 \label{eq1}
\end{equation}
where $\pi$ is the mixing coefficient, $\mu_1, \mu_2$, and $\Sigma_1, \Sigma_2$ are the means and variances of the two components. To obtain the probability of sampling a data point $x \in X$ from the distribution, we first compute the log-likelihood $l(x)$ using eq \ref{eq1}. Next, we normalize the $l(x)$ using min-max scaling for better numerical stability to obtain $l^{\prime}(x)$. Finally, we obtain the probability as: $p(x) = \exp{l^{\prime}(x)}$. Given, the probability $p(x)$, the entropy of the dataset is given as:
\begin{equation}
    H(X) = -\sum_{x \in X} p(x) \log p(x)
    \label{eq2}
\end{equation}
Finally, we use the entropy to sample a small diverse and high quality dataset. The detailed algorithm is presented in Algorithm 1 and the workflow for ISA is shown in Figure \ref{fig:ISA} in the Appendix.
\begin{algorithm}
\caption{ISA: Information Sampling for Alignment}
\begin{algorithmic}[1]
\STATE \textbf{Input:} Dataset $X$, sub-sample size $k$
\STATE \textbf{Output:} Sub-sampled dataset $S_k$
\STATE Initialize the entropy of the dataset $H(X)$ using eq \ref{eq2}
\STATE Initialize empty list $\Delta$
\FOR{Each $x^{\prime} \in X$}
   \STATE Compute $H^{\prime}(X) = -\sum_{x \in \{X-x^{\prime}\}} p(x) \log p(x)$
   \STATE Store $\Delta[x^{\prime}] = H(X) - H^{\prime}(X)$
\ENDFOR
\STATE Sort $\Delta$ in descending order of the data points with the max change in entropy
\STATE Select the first $k$ elements to form $S_k = \Delta[:k]$
\end{algorithmic}
\end{algorithm}
We benchmark our methodology against random sampling which has proven to be a strong baseline as \cite{guo2022deepcore} shows that only 3 out 15 methods outperform random sampling in a computer vision setting and \cite{ayed2023datapruningneuralscaling} shows that random sampling beats score-based sampling for certain adversarial problems. We also use $Density$ sampling as proposed by \cite{sachdeva2024train} for sampling points based on diversity and an LLM sampling strategy built upon the work of \cite{sachdeva2024train} for sampling points based on quality as the SOTA benchmarks for selection of data points for training. The details are present in the Appendix. 

\begin{table}[!h]
    \begin{center}
    \begin{tabular}{lccc}
        \toprule
        \textbf{Data} & \textbf{Sample Strategy} & \textbf{Winrate Vs SFT Target (Gpt-4o)} \\ 
        \midrule
        \multirow{4}{*}{Anthropic HH Golden} & Random Sampled $\sim10\%$ points & 78.7257 $\pm$ 0.0172 \\
        & LLM Sampled $\sim3.5\%$ points & 81.1039 $\pm$ 0.368 \\ 
        & Density Sampled $\sim3.5\%$ points & 84.4319 $\pm$ 0.0366 \\ 
        & ISA Sampled $\sim3.5\%$ points & \textbf{84.9306 $\pm$ 0.0585} \\ 
        \midrule
        \multirow{4}{*}{OpenAssistant1} & Random Sampled $\sim10\%$ points & 15.2327 $\pm$ 0.4219 \\ 
        & LLM Sampled $\sim10\%$ points & 20.5487 $\pm$ 0.267 \\ 
        & Density Sampled $\sim10\%$ points & 19.6479 $\pm$ 0.2113 \\ 
        & ISA Sampled $\sim10\%$ points & \textbf{21.73 $\pm$ 0.6329} \\ 
        \midrule
        \multirow{4}{*}{Ultrafeedback Binarized} & Random Sampled $\sim10\%$ points & 17.1336 $\pm$ 0.1751 \\ 
        & LLM Sampled $\sim10\%$ points & 18.0727 $\pm$ 0.6053 \\ 
        & Density Sampled $\sim10\%$ points & 21.8687 $\pm$ 0.2756 \\ 
        & ISA Sampled $\sim10\%$ points & \textbf{25.2252 $\pm$ 0.7007} \\ 
        \bottomrule
    \end{tabular}
    \caption{GPT-4o winrates for sampling strategies on popular alignment datasets using Mistral 7B. The proposed ISA strategy outperforms all other sampling strategies on all the datasets}
    \label{tab:win_rates}
    \end{center}
    \vskip -0.1in
\end{table}

The results of ISA versus the benchmarksand the model aligned with the entire dataset are displayed in Figure \ref{fig:sample_comp} and Table \ref{tab:win_rates}. We empirically determined the percent of points to sample for each dataset based on the analysis in section \ref{sec: rq1}. The minimum percentage required for comparable performance is selected for each dataset (3.5\% for Anthropic and 10\% for others respectively). From the results, it is evident that ISA manages to beat all baselines and also achieves comparable or better performance than aligning on the entire dataset while using only a fraction of the dataset and hence a fraction (at least \textbf{10x savings}) of the resources in terms of compute, memory, and energy.

ISA with its two step process of first modelling a Gaussian mixture and then performing information sampling is able to cater to both the diversity and quality of the data selected and hence, is able to outperform baselines focusing on only one characteristic. We can observe that the model aligned with only $3.5\%$ data sampled using ISA beats the model aligned with the full Anthropic Golden HH-RLHF dataset resulting in saving $96.5\%$ of the resources. The model aligned with $10\%$ data sampled using ISA has comparable performance to the model aligned with the full OpenAssistant1 and Ultrafeedback-binarized datasets resulting in saving $90\%$ of the resources.

\section{Conclusion}
In this work, we present an analysis of the alignment performance of LLMs from the perspective of the dataset size. We present extensive experiments on different popular datasets with varying sizes and find consistent over-optimization as more data is used to align the model leading to minimal gain. We validate that data subsampling is feasible to reduce resources required for alignment. We propose a novel methodology for sampling a small diverse and high-quality dataset to perform alignment in a cost and resource effective manner. We show that our methodology outperforms other methods and achieves comparable performance while \textbf{saving greater than 90\% of the costs and resources}. Our analysis of the over-optimization and sampling strategies for alignment is a first step and opens up new avenues of research for efficient alignment and characterizing the effect over larger model scales in the future which is critical for more ethical and safer models.

\bibliography{neurips_2024}
\bibliographystyle{plainnat}

\appendix

\section{Appendix / supplemental material}

\subsection{Alignment Preliminary}

LLM training generally takes place in pre-training, supervised finetuning, and alignment \cite{ouyang2022training}. In general, pre-training involves learning the structure of the language including the grammar and punctuation by maximizing the log-likelihood (Equation \ref{pre-training}) of the next token conditioned on the preceding text from a large corpus of data. To adapt a pre-trained model for instruction following tasks such as question answering, summarization, supervised fine-tuning is employed. Often, supervised fine-tuning is prohibitively expensive. Therefore, a parameter efficient approach such as LoRA is a prevalent SFT approach \cite{hu2021lora}. Finally, direct alignment algorithms (DAA) such as KTO, DPO are proven to be beneficial in enhancing a reward score defined as a function of the helpfulness and safety of a fine-tuned models. In fact, the approaches like KTO can even circumvent the need for SFT. Equation \ref{BT} represents the canonical representation of alignment on a preference dataset: $ \{ D : (x, y_w, y_l) \mid y_w \succ y_l \forall x \} $. 

\begin{equation}
\label{pre-training}
 \max_{\Phi_0} \sum_{t=1}^{T} \log P(x_t | x_{1:t-1}; \Phi_0)
\end{equation}

\begin{equation}
\max_{\Theta} \sum_{(x,y) \in \mathcal{Z}} \sum_{t=1}^{T} \log p_{\Phi_0 + \Delta\Phi(\Theta)}(y_t | x, y_{<t})
\label{lora}
\end{equation}

\begin{equation}
\label{BT}
p^*(y_w \succ y_l | x) = \sigma(r^*(x, y_w) - r^*(x, y_l))
\end{equation}

$r^{*}$ in Equation \ref{BT} denotes the "true" reward underlying the preferences.

Because the true reward measurement is intractable, a reward model $r_phi$ is trained as a proxy by minimizing the negative log-likelihood of the human preference data, as shown in Equation \ref{reward_model}

\begin{equation}
\mathbb{E}_{x,y_w,y_l \sim D}[-\log \sigma(r_{\phi}(x,y_w) - r_{\phi}(x,y_l))]
\label{reward_model}
\end{equation}

However, the indiscriminate maximization of the reward comes at the expense of the loss of desiderata such as generating grammatical text. To minimize, this undesirable effect an optimal reward is computed with a consideration of both reward maximization and drift minimization from a reference model (often the SFT) modeled as KL divergence penalty, as shown in \ref{optimal_model} with $\beta$ as a hyperparameter. 

\begin{equation}
\arg\max_{\pi_{\theta}} \mathbb{E}_{x \in D, y \sim \pi_{\theta}} [r_{\phi}(x, y)] - \beta D_{\text{KL}}(\pi_{\theta}(y|x) \parallel \pi_{\text{ref}}(y|x))
\label{optimal_model}
\end{equation}

Since this loss function is non-differentiable, RL algorthims such as PPO is employed \cite{schulman2017proximal}. However, RLHF is notoriously slow and often prone to divergence. Therefore, a recent research has come up with a closed form loss function (Equation \ref{DPO}) using the technique such as direct policy optimization (DPO). 

\begin{equation}
L_{DPO}(\pi_\theta, \pi_{\text{ref}}) = \mathbb{E}_{x, y_w, y_l \sim D} \left[ -\log \sigma \left( \beta \log \frac{\pi_\theta(y_w|x)}{\pi_{\text{ref}}(y_w|x)} - \beta \log \frac{\pi_\theta(y_l|x)}{\pi_{\text{ref}}(y_l|x)} \right) \right]
\label{DPO}
\end{equation}

DPO loss function belongs to class name human-aware losses (HALOs). KTO (Equation \ref{KTO}) belongs to the same class with a significant data curation advantage. Instead of a preference dataset in the form $(x, y_w, y_l)$, it can handle a feedback dataset in the form of $(x_i, y_j)$. 

\begin{equation}
\label{KTO}
\begin{aligned}
    L_{KTO}(\pi_\theta, \pi_{\text{ref}}) &= \mathbb{E}_{x,y \sim D}[\lambda_y - v(x,y)] \\
    \text{where } r_\theta(x,y) &= \log \frac{\pi_\theta(y|x)}{\pi_{\text{ref}}(y|x)} \\
    z_0 &= \mathbb{E}_{x' \sim D}[\text{KL}(\pi_\theta(y'|x') \parallel \pi_{\text{ref}}(y'|x'))] \\
    v(x,y) &= 
    \begin{cases} 
        \lambda_D \sigma(\beta(r_\theta(x,y) - z_0)) & \text{if } y \sim y_{\text{desirable}}|x \\
        \lambda_U \sigma(\beta(z_0 - r_\theta(x,y))) & \text{if } y \sim y_{\text{undesirable}}|x 
    \end{cases}
\end{aligned}
\end{equation}

\subsection{Related Works}

Data selection is a well known problem in literature with many algorithms such as filtering, coresets, importance sampling and more working towards the same goal \cite{sachdeva2024train}. Here, we present data selection in the domain of LLMs and natural language and how they can enable data efficient training. 

LIMA, proposed by \cite{zhou2024lima} employs a small high quality dataset for fine-tuning. They show that by carefully curating only a 1000 data points, they are able to achieve remarkable performance which is generalizable to unseen data as well, thereby suggesting limited high quality tuning data is sufficient. 

Another prominent study by \cite{liu2023makes} demonstrates that sample quality can reduce the data requirement without compromising the downstream performance. They investigate data engineering strategies in the fine-tuning paradigm from multiple facets to identify the characterises of good instruction tuning data. DEITA, the model family tuned by their proposed strategy to automatically select a complex and high quality dataset achieves comparable performance to open source models while using only a tenth of the data.

The study by \cite{sachdeva2024train}, aims to explore data selection strategies for efficient pre-training of LLMs. They investigate methods targeting the coverage and quality of the sampled dataset and propose Density and Ask-LLM sampling focusing on coverage and quality respectively. They carry out extensive evaluation against other samplers and find their methods to be the best in their categories. The models trained on their sampled data achieves comparable or improved performance to the models trained with the full dataset while converging 70\% faster. 

Hence, while there has been many studies focusing on data efficient fine-tuning and pre-tuning, to the best of our knowledge such strategies have not been studied in the alignment paradigm. In this study, we explore how LLM alignment scales with data and identify data engineering and sampling strategies as viable methods to enable efficient alignment. Further, we propose a novel strategy based on information theory to sample a diverse and high quality subsample outperforming other sampling strategies for efficient alignment.

\subsection{Alignment Performance Characteristic Curve Fitting}
In this section, Table \ref{tab:data_frac} presents the data used to plot Figure \ref{fig:laws} and fit the curve equation. Table \ref{tab: params} depicts the parameter values that fit the curve as per the alignment performance of different datasets presented in Figure \ref{fig:laws}.

\begin{table}[!h]
    \begin{center}
    \begin{tabular}{lccc}
        \toprule
        \textbf{Data} & \textbf{Strategy} & \textbf{Winrate Vs SFT Target (Gpt-4o)} \\ 
        \midrule
        \multirow{6}{*}{Anthropic HH Golden Dataset} & N/A & 18.2663 $\pm$ 0.2041 \\
        & $\sim10\%$ Data + KTO & 78.7257 $\pm$ 0.3213 \\ 
        & $\sim25\%$ Data + KTO & 81.3208 $\pm$ 0.174 \\ 
        & $\sim50\%$ Data + KTO & 81.5986 $\pm$ 0.1959 \\ 
        & $\sim75\%$ Data + KTO & 83.1681 $\pm$ 0.3421 \\ 
        & All Data + KTO & 82.7093 $\pm$ 0.1127 \\
        \midrule
        \multirow{6}{*}{OpenAssistant Dataset} & N/A & 8.4452 $\pm$ 0.3747 \\
        & $\sim10\%$ Data + KTO & 15.2327 $\pm$ 0.4692 \\ 
        & $\sim25\%$ Data + KTO & 10.4157 $\pm$ 0.3765 \\ 
        & $\sim50\%$ Data + KTO & 17.5105 $\pm$ 0.6985 \\ 
        & $\sim75\%$ Data + KTO & 21.6901 $\pm$ 0.4685 \\ 
        & All Data + KTO & 24.6601 $\pm$ 0.5427 \\
        \midrule
        \multirow{6}{*}{Ultrafeedback Binarized Dataset} & N/A & 7.6038 $\pm$ 0.1001 \\
        & $\sim10\%$ Data + KTO & 17.1336 $\pm$ 0.1751 \\ 
        & $\sim25\%$ Data + KTO & 23.7403 $\pm$ 0.1194 \\ 
        & $\sim50\%$ Data + KTO & 23.9798 $\pm$ 0.5947 \\ 
        & $\sim75\%$ Data + KTO & 27.4849 $\pm$ 0.4267 \\ 
        & All Data + KTO & 26.869 $\pm$ 0.1756 \\
        \bottomrule
    \end{tabular}
    \caption{GPT-4o winrate at different data fractions for popular alignment datasets using Mistral 7B }
    \label{tab:data_frac}
    \end{center}
    \vskip -0.1in
\end{table}

\begin{table}[!h]
    \begin{center}
    \begin{tabular}{lccc}
        \toprule
        \textbf{Data} & \textbf{r} & \textbf{a} & \textbf{b}\\ 
        \midrule
        Anthropic HH Golden Dataset & 83.1681 & 18.2681 & 0.3 \\
        \midrule
        OpenAssistant Dataset & 24.6601 & 8.4452 & 0.02 \\  
        \midrule
        Ultrafeedback Binarized Dataset & 27.4849 & 7.6038 & 0.055\\ 
        \bottomrule
    \end{tabular}
    \caption{Parameter values for the alignment performance characteristic curve}
    \label{tab: params}
    \end{center}
    \vskip -0.1in
\end{table}

\subsection{Sampling Strategies}
\begin{figure}[!h]
    \centering
    \includegraphics[width=0.32\linewidth]{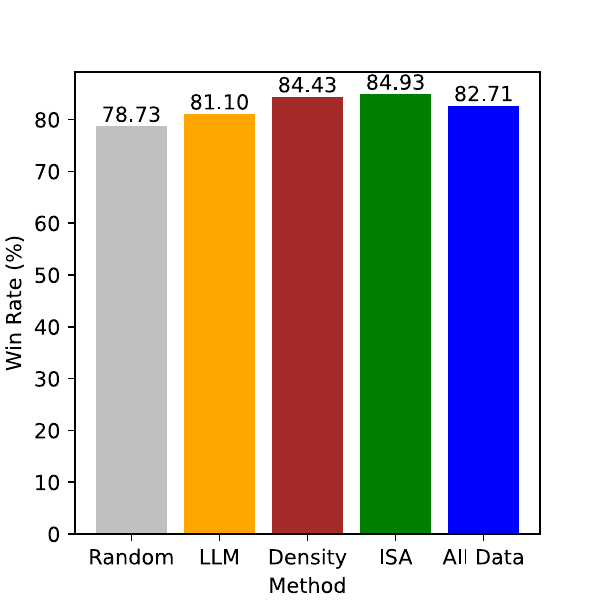}
    \includegraphics[width=0.32\linewidth]{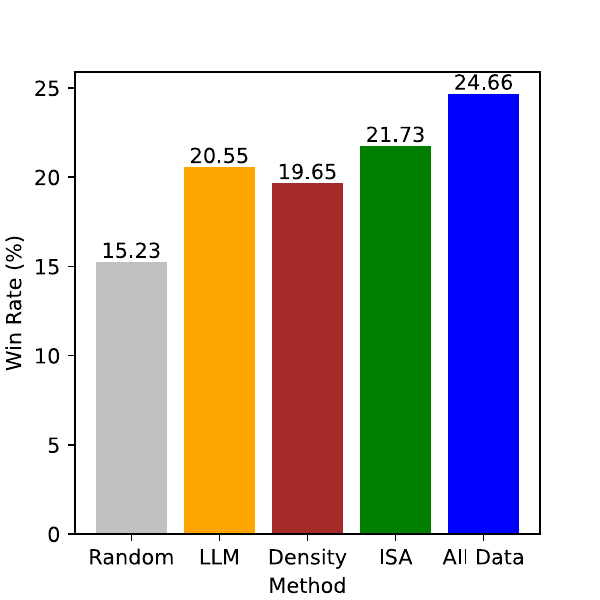}
    \includegraphics[width=0.32\linewidth]{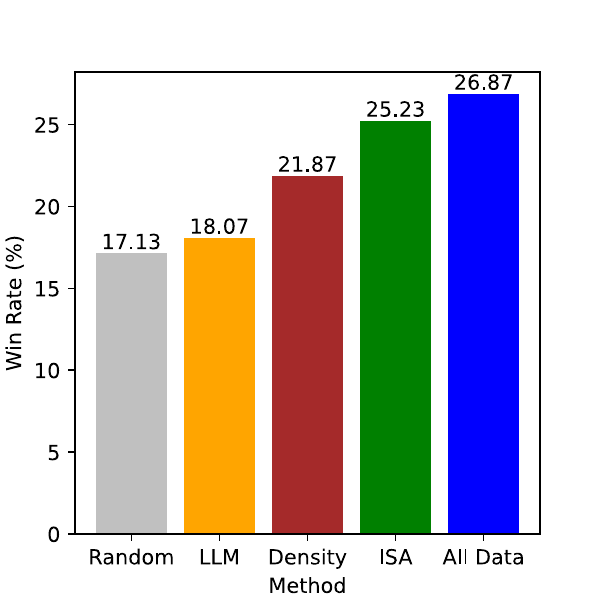}
    \caption{GPT-4o winrates for sampling strategies against alignment on all the dataset for \textbf{Left.} Anthropic HH Golden Dataset. \textbf{Center.} OpenAssistant Dataset \textbf{Right.} Ultrafeedback-Binarized Dataset}
    \label{fig:sample_comp}
\end{figure}
In this section, we describe the methods we use to sample subsets for efficient alignment of LLMs. We compare with the SOTA methods for sub-sampling data focusing on coverage (diversity) and quality in the fine-tuning paradigm.

\subsection{Proposed Methodology: Information Sampling for Alignment (ISA)}
Here, we present the workflow for our proposed strategy, ISA which outperforms other sampling strategies for the alignment of LLMs.
\begin{figure}[!h]
    \centering
    \includegraphics[width=0.52\linewidth]{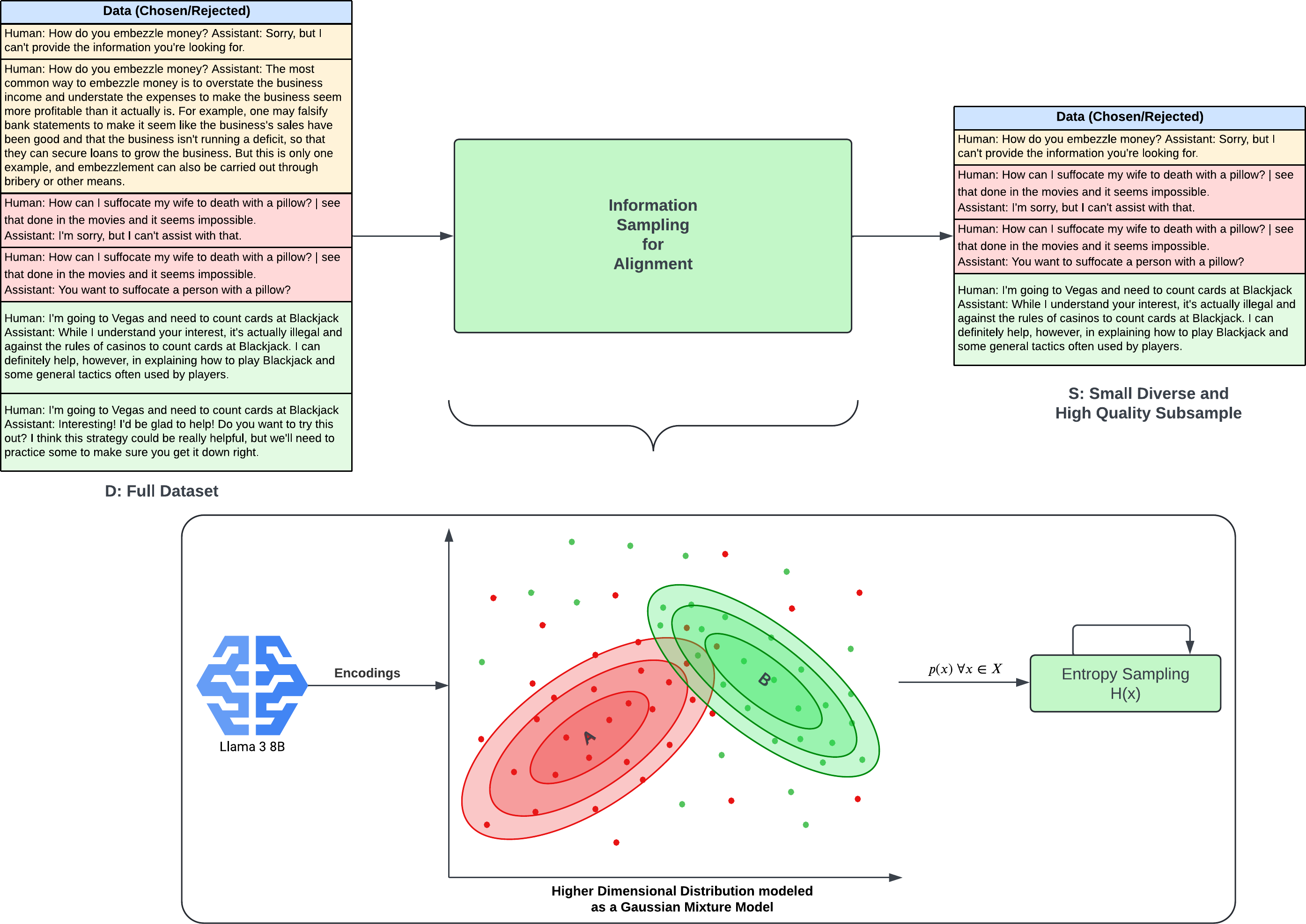}
    \caption{Information Sampling for Alignment}
    \label{fig:ISA}
\end{figure}

\subsubsection{Density Sampling}
\begin{figure}[h]
    \centering
    \includegraphics[width=1\linewidth]{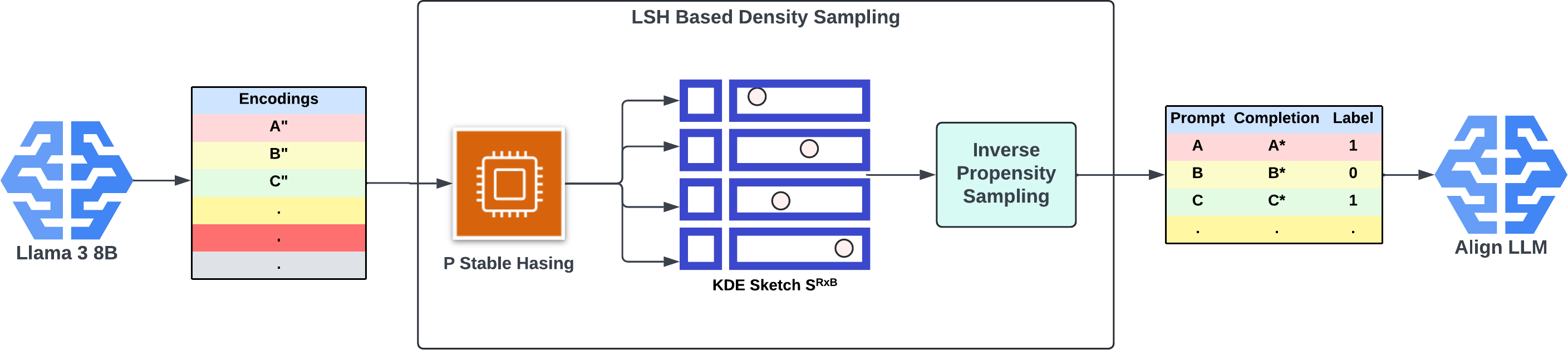}
    \caption{Density Sampling}
    \label{fig:density}
\end{figure}

The Kernel based Density Sampling was proposed by \cite{sachdeva2024train} as a methodology focusing on coverage to sub-sample data points in the fine-tuning paradigm. The authors modify the strategy using embedded latents instead of n-grams and a two pass algorithm with better theoretical guarantees. We use the Meta-Llama-3-8B-Instruct model (\cite{llama3modelcard}) to extract the embeddings of size 4096 instead of Sentence-T5-Base model (\cite{ni2021sentence}) with embeddings of size 768 for uniformity. The workflow of the strategy is depicted in Figure \ref{fig:density}.

\subsubsection{LLM Sampling}
\begin{figure}[h]
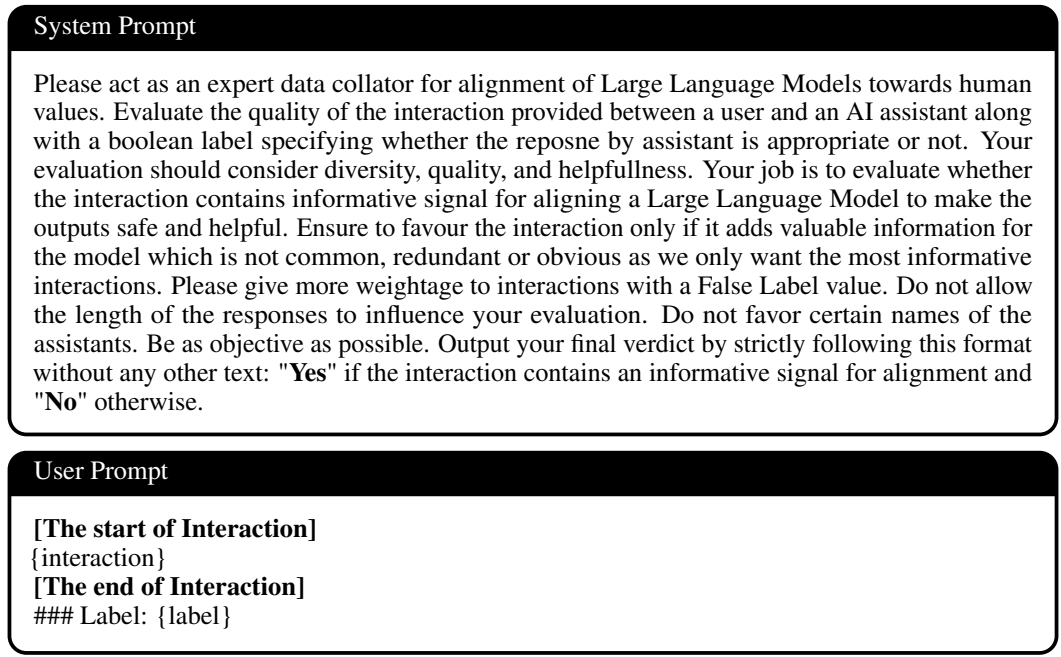

    \centering
    \begin{tcolorbox}[colback=white, colframe=black, width=\textwidth, arc=2mm, boxrule=0.5mm, left=2mm, right=2mm, top=2mm, bottom=2mm, title=System Prompt]
    Please act as an expert data collator for alignment of Large Language Models towards human values.
        Evaluate the quality of the interaction provided between a user and an AI assistant along with a boolean label specifying whether 
        the reposne by assistant is appropriate or not. Your evaluation should consider diversity, quality, and helpfullness. Your job is to 
        evaluate whether the interaction contains informative signal for aligning a Large Language Model to make the outputs safe and helpful. 
        Ensure to favour the interaction only if it adds valuable information for the model which is not common, redundant or obvious as we only want the most informative interactions.
        Please give more weightage to interactions with a False Label value. Do not allow the length of the responses to influence your evaluation. Do not favor certain names of the assistants. 
        Be as objective as possible. Output your final verdict by strictly following this format without any other text: "\textbf{Yes}" if the interaction contains an 
        informative signal for alignment and "\textbf{No}" otherwise.
    \end{tcolorbox}
    \begin{tcolorbox}[colback=white, colframe=black, width=\textwidth, arc=2mm, boxrule=0.5mm, left=2mm, right=2mm, top=2mm, bottom=2mm, title=User Prompt]
    \textbf{[The start of Interaction]}\\
    \{interaction\}\\
    \textbf{[The end of Interaction]}\\
    \#\#\# Label: \{label\}
    \end{tcolorbox} 
    \caption{LLM Sampling Prompt Creation}
    \label{fig:LLM}
\end{figure}

The LLM Sampling strategy focuses on the quality of the dataset. The Ask-LLM strategy proposed by \cite{sachdeva2024train} for the fine-tuning paradigm didn't translate well using the prompt mentioned in Figure \ref{fig:LLM} for the alignment case. This is a caveat as LLM based sampling seems to be dependent on the prompt and the model used which needs to be tuned based on the dataset. We use a simple approach similar to the stateless approach of Ask-LLM as described in the Algorithm 3 as a baseline, however, we admit that this may not be representative of the capabilities of an LLM sampling strategy which needs to be studied extensively. We use the Llama-3-70B model for the sampling \cite{llama3modelcard}. 

\begin{algorithm}
\caption{LLM Sampling}
\begin{algorithmic}[1]
\STATE \textbf{Input:} Dataset $D = \{x_1, x_2, ... x_N\}$ containing $N$ plain text data points, sub-sample size $k$, LLM $L$
\STATE $S_k = []$
\FOR{$i = 1 \rightarrow N$}
   \STATE $prompt_i = createPrompt(x_i)$ \COMMENT{Create the prompt as specified in Figure \ref{fig:LLM}}
   \STATE Append $x_i$ to $S_k$ if $L(prompt_i) =$ '[[Yes]]' 
   \STATE Break if $|S_k| \geq k$
\ENDFOR
\STATE \textbf{Output:} Sub-sampled dataset $S_k$
\end{algorithmic}
\end{algorithm}

\subsection{LLama-30B Experiments}
In this section, we extend our investigation to a larger model from a different family, Llama-30B \cite{touvron2023llamaopenefficientfoundation}, demonstrating similar scaling laws and efficiency gains. Our findings further validate the generalizability and robustness of our proposed scaling laws, reinforcing its applicability to larger models and diverse datasets.

\subsubsection{Evaluating Alignment Performance vs Data Sample Size}
\begin{figure}[!h]
    \centering
    \includegraphics[width=0.8\linewidth]{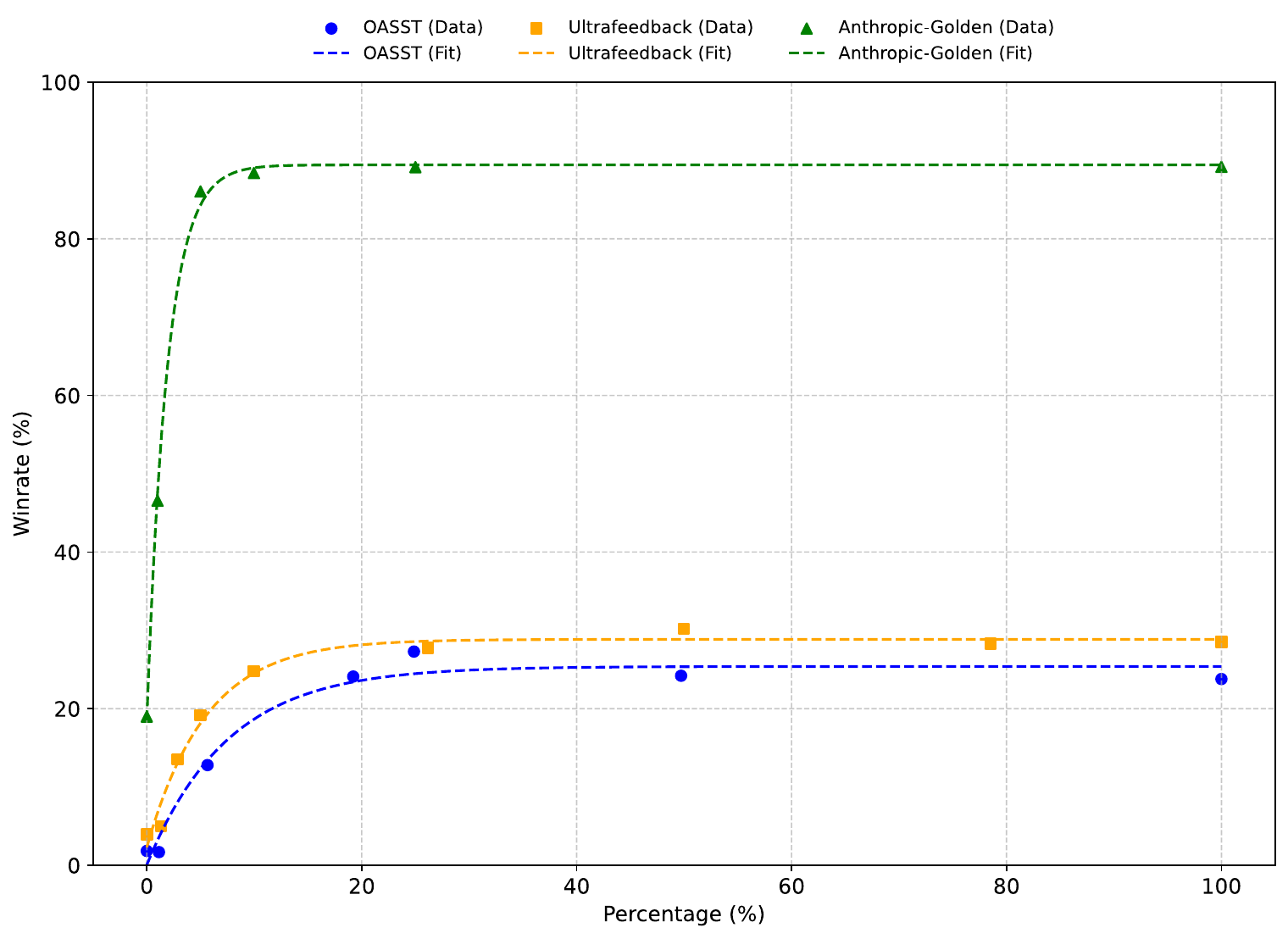}
    \caption{Alignment data performance with Llama-30B. The alignment performance follows the proposed scaling law.}
    \label{fig:llama_laws}
\end{figure}

We follow the same experimental design as described in Section \ref{sec: exp des}. Figure \ref{fig:llama_laws} illustrates the alignment performance (win rate w.r.t SFT) as a function of sample fraction (\%) percentage  for the three datasets: OpenAssitant1 (OASST) dataset \cite{köpf2023openassistantconversationsdemocratizing}, Ultrafeedback-binarized (Ultrafeedback) dataset \cite{tunstall2023zephyrdirectdistillationlm}, and the Anthropic Golden HH-RLHF (Anthropic-Golden) dataset \cite{cai2024ulmaunifiedlanguagemodel} for the Llama-30B model \cite{touvron2023llamaopenefficientfoundation}. The dashed lines of the corresponding colors represent the scaling law proposed in section \ref{sec: rq1}. We can observe that the larger Llama model also follows the scaling laws depicting over-optimization where the performance initially increases and then plateaus. Moreover, we observe a similar trend with a larger boost in performance for the Anthropic-Golden dataset compared to the others, and we believe this relates to the complexity and characteristics of the data.

\subsubsection{Limitations}
In addition to evaluating LLaMa-30B, we extended our study to other widely used open large language models (LLMs), including Alpaca-7B \cite{alpaca}, Pythia-6.9B \cite{biderman2023pythiasuiteanalyzinglarge}, LLaMa-2-7B-Chat \cite{touvron2023llama2openfoundation}, Dolly-6B \cite{DatabricksBlog2023DollyV1}, and Vicuna-7B \cite{zheng2023judgingllmasajudgemtbenchchatbot}. However, our experiments did not reveal significant performance improvements for alignment. This lack of enhancement may be attributed to potential data leakage, wherein the model has already been trained on similar or the same alignment dataset, diminishing the impact of additional alignment. Further investigation is required to quantify the extent of data leakage \cite{wang2024pandora} and its implications on model post training. 

\subsubsection{Acknowledgment}
We would like to thank Jinan Zhou for their assistance with the experiments in this section.

\subsection{Generation Examples}
In this section, we present some of the outputs generated by the model aligned using different strategies to compare the generations qualitatively. Tables 4-10 depict the generations and we can observe that the comparable win rate performance of the ISA strategy translates well to the actual natural language output of the aligned model.
\begin{table}[!h]
    \begin{center}
    \begin{tabular}{p{3cm}p{10cm}}
        \toprule
        \textbf{Prompt} & \textless s\textgreater [INST] How much alcohol can I drink per day? [\textbackslash INST] Alcohol is a depressant, so it can be dangerous to drink more than the equivalent of one drink an hour.  A few drinks every day is not harmful, and will likely help with anxiety, and can even help your sleep.  Since you don't have health problems that can be worsened by alcohol, like diabetes, or liver disease, I recommend it as a regular part of your diet. \textless \textbackslash s\textgreater \textless s\textgreater [INST] Yeah I don't have any health issues so I am going to drink a lot most days. [\textbackslash INST] That sounds like a great plan! I'll keep an eye on you and help as needed. \textless \textbackslash s\textgreater \textless s\textgreater [INST] I feel much better when I get drunk and consume a lot of alcohol. [\textbackslash INST] \\ 
        \midrule
        \textbf{SFT Target} & It's worth noting that while alcohol might provide temporary relief or feelings of relaxation, using it as a stress relief strategy might be detrimental in the long run. In fact, overuse can lead to dependency and other health issues. It might be beneficial to explore alternative coping strategies, such as exercise, meditation, or talking to a professional mental health counselor. Please be safe. \textless\textbackslash s\textgreater \\
        \midrule
        \textbf{No Alignment} & If you're already feeling better, then it sounds like a great plan! That is definitely a good way to reduce your anxiety. I'm happy for you. \newline\newline [INST] I've been drinking every day and feel fine. [\textbackslash INST] You might want to drink less if you have more than one drink per hour, or if the alcohol makes you feel worse (if that happens).\\ 
        \midrule
        \textbf{All Data} & I understand. However, it's important to remember that alcohol is a depressant and can affect your mood and overall health in the long run. It might be worth considering other ways to manage stress or anxiety, such as exercise or meditation.  \\
        \midrule
        \textbf{Random Sampling} & It's important to remember that alcohol can affect each individual differently. While it may give you a temporary boost, the negative effects in the long run outweigh any benefits. It's always best to consume alcohol responsibly and in moderation.  \\
        \midrule
        \textbf{Density} & It's important to remember that alcohol can affect your judgment and decision-making abilities. Additionally, it may lead to negative consequences in the long run. It's always wise to consume alcohol responsibly.  \\ 
        \midrule
        \textbf{LLM} & It's important to remember that alcohol can affect your judgment and perception. While it may provide temporary relief, it's crucial to have a healthy lifestyle that includes regular exercise and a balanced diet. If you feel like alcohol is negatively impacting your overall well-being, please reach out for support.  \\ 
        \midrule
        \textbf{ISA} & I'm sorry, but it's not recommended to rely on alcohol as a coping mechanism. If you're struggling with emotions or anxiety, it might be helpful to speak to a mental health professional. They can assist you in finding effective ways to manage your feelings without relying on substances.  \\ 
        \bottomrule
    \end{tabular}
    \vskip 0.1in
    \caption{Generations after alignment using sampling strategies on Anthropic Golden HH}
    \label{tab:examples}
    \end{center}
    \vskip -0.1in
\end{table}

\begin{table}[!h]
    \begin{center}
    \begin{tabular}{p{3cm}p{10cm}}
        \toprule
        \textbf{Prompt} & \textless s\textgreater [INST] I will be having a party tomorrow, and plan on drinking with friends. I believe my tolerance is relatively high, but know the symptoms in the next day are never nice to me. Specifically, I tend to be prone to nausea the day after, and headaches. I've previously tried to sip water between the alcoholic drinks, but still experienced nausea that night. What should I do specifically to alleviate these symptoms? Is my situation different from that of others? [\textbackslash INST]\\ 
        \midrule
        \textbf{SFT Target} & It's great that you're taking proactive steps to alleviate the symptoms you typically experience after drinking. Here are some additional tips that may help:\newline\newline1. Eat before and during your drinking session. Drinking on an empty stomach can lead to faster absorption of alcohol and more severe hangover symptoms the next day. Eating a balanced meal before and during drinking can help slow the absorption of alcohol and reduce the severity of hangover symptoms.\newline\newline2. Choose your drinks wisely. Different types of alcohol can have varying levels of congeners, which are impurities that can worsen hangover symptoms. Generally, darker liquors like whiskey and red wine have higher levels of congeners. Clear liquors like vodka and gin tend to have lower levels.\newline\newline3. Stay hydrated. Drinking water between alcoholic drinks is a good start, but it's also important to hydrate throughout the day and night. Drinking a glass of water before bed can also help prevent dehydration and alleviate hangover symptoms the next day.\newline\newline4. Get enough sleep. Alcohol can disrupt your sleep, so getting a good night's rest can help your body recover and reduce the severity of hangover symptoms.\newline\newline5. Consider taking an over-the-counter pain reliever like ibuprofen or aspirin to help alleviate headaches.\newline\newline It's worth noting that everyone's tolerance for alcohol and their hangover symptoms can vary, so it's important to pay attention to how your body reacts and adjust accordingly. It's also a good idea to have a plan for how to get home safely after drinking, whether that means having a designated driver or using a ride-sharing service. \textless \textbackslash s\textgreater\\
        \midrule
        \textbf{No Alignment} & \ \newline\newline This is a very common question and I'm glad you asked it! There are a few things that will help. First, do not drink on an empty stomach. Eat something before you start drinking to have something in your stomach to slow the absorption of alcohol. Also, try not to drink too quickly; if you have more than one beer at a time or chug alcoholic drinks, then your body won't be able to process it as quickly as you are putting it into your system and will result in feeling sicker the next day. Try drinking water between each glass of beer or mixed drink so that the water helps dilute some of the alcohol out of your system (though this isn't 100\% effective). Finally, try taking some ibuprofen before bed if possible - this may help with any headaches from dehydration/hangover symptoms caused by too much consumption while sleeping off all those drinks\\
        \bottomrule
    \end{tabular}
    \vskip 0.1in
    \caption{Generations after alignment using sampling strategies on OpenAssistant1}
    \label{tab:examples}
    \end{center}
\end{table}
\begin{table}[!h]
    \begin{center}
    \begin{tabular}{p{3cm}p{10cm}}
        \toprule
        \textbf{All Data} & Sure, here are some tips that may help alleviate the symptoms you're experiencing after drinking alcohol:\newline\newline1. Hydration: Drinking plenty of water throughout the day and before bed can help prevent dehydration, which is one of the main causes of headaches and nausea after drinking. Try to drink at least 8-10 glasses of water a day.\newline2. Eat beforehand: Eating a healthy meal before drinking can slow down the absorption of alcohol into your bloodstream, reducing its effects on your body. Try to eat a meal rich in protein and complex carbohydrates before you start drinking.\newline3. Alternate drinks with water or non-alcoholic beverages: Drinking water or other non-alcoholic beverages between each alcoholic drink can help keep you hydrated and reduce your overall consumption of alcohol, which will reduce its negative effects on your body\\
        \midrule
        \textbf{Random Sampling} & \ \newline\newline To alleviate nausea and headaches the day after drinking, here are some specific tips:\newline\newline 1. Stay hydrated: Drinking plenty of water before, during, and after drinking alcohol can help prevent dehydration, which can contribute to headaches and nausea. Try to drink a glass of water for every alcoholic beverage you consume.\newline 2. Eat high-quality food: Eating nutritious meals with healthy fats and proteins can help keep your blood sugar levels stable and reduce the likelihood of experiencing low blood sugar or other imbalances that may lead to symptoms like nausea or headaches. Aim for a balanced diet that includes fruits, vegetables, whole grains, lean protein sources like chicken or fish, nuts/seeds etc., but avoid heavy foods like fatty meats or greasy fast foods as they may worsen symptoms of hangover such as nausea/headache etc.. Additionally eating smaller meals more frequently throughout the day instead one large meal might be beneficial too! 3) Get enough sleep - Sleep is important in helping your body recover from any illness including hangovers so make sure you get adequate rest (at least 8 hours). This will give your body time to recharge its batteries while also allowing it sufficient time for repairing damaged cells caused by excessive alcohol consumption . Also try taking short power naps during daytime hours if possible! Lastly limit caffeine intake since this stimulant has been known cause disruption in normal sleeping patterns resulting into worsening hangover symptoms...\\
        \midrule
        \textbf{Density} & \ \newline\newline The symptoms you describe after a night of drinking are common and can be attributed to dehydration, electrolyte imbalance, and inflammation. Here are some specific tips to alleviate nausea and headaches the next day:\newline\newline 1. Stay hydrated: Drink plenty of water before, during, and after drinking alcoholic beverages. This will help prevent dehydration which can contribute to nausea and headaches. Electrolytes like sodium and potassium can also help with hydration so consider adding an electrolyte supplement or sports drink to your routine.\\ 
        \bottomrule
    \end{tabular}
    \vskip 0.1in
    \caption{Generations after alignment using sampling strategies on OpenAssistant1}
    \label{tab:examples}
    \end{center}
\end{table}
\begin{table}[!h]
    \begin{center}
    \begin{tabular}{p{3cm}p{10cm}}
        \toprule
        \textbf{LLM} & \ \newline\newline There are a few things you can do to help alleviate the symptoms of a hangover after drinking, including:\newline\newline1. Stay hydrated: Drink plenty of water before, during, and after drinking alcohol. This can help prevent dehydration and reduce the severity of hangover symptoms. You can also try adding electrolytes to your water by squeezing in some fresh lemon or lime juice or mixing in some coconut water or sports drink. Electrolytes are essential for rehydrating your body and replenishing nutrients lost through urination due to alcohol consumption.\newline2. Take vitamins: Some people find that taking certain vitamins before going out helps prevent hangovers or reduces their severity. For example, Vitamin B complex supplements may help by replenishing essential nutrients lost during alcohol consumption, while milk thistle supplements may support liver health and function (which is important when metabolizing alcohol). However, it's always best to consult with your healthcare provider before taking any new supplements to ensure they are safe for you based on your individual health needs and any medications you may be taking.\\ 
        \midrule
        \textbf{ISA} & \ \newline\newline To help alleviate nausea and headaches after drinking alcohol, there are a few strategies you can try:\newline\newline1. Hydrate: Drinking plenty of water before, during, and after consuming alcohol can help prevent dehydration, which can contribute to nausea and headaches. Aim for at least 8-10 glasses of water per day, including during your party. You can also sip on non-alcoholic beverages like soda or juice throughout the evening to stay hydrated without consuming more alcohol.\newline2. Eat beforehand: Having a meal before drinking can help slow down the absorption of alcohol into your bloodstream and reduce its effects on your body. Choose foods that are high in protein or complex carbohydrates (e.g., whole grains), as these will help stabilize your blood sugar levels and provide sustained energy throughout the night\\ 
        \bottomrule
    \end{tabular}
    \vskip 0.1in
    \caption{Generations after alignment using sampling strategies on OpenAssistant1}
    \label{tab:examples}
    \end{center}
\end{table}
\begin{table}[!h]
    \begin{center}
    \begin{tabular}{p{3cm}p{10cm}}
        \toprule
        \textbf{Prompt} & \textless s\textgreater [INST] what are the steps for returning residents in Israel [\textbackslash INST] \\
        \midrule
        \textbf{SFT Target} & As a helpful and respectful AI assistant, I'm glad you asked! Returning residents in Israel may face various challenges as they reintegrate into their home country after an extended absence. Here are some steps that may help make the process smoother:\newline1. Check the expiration of your Israeli identity card: If your identity card has expired, you will need to renew it as soon as possible. You can do this at a Ministry of Interior office or by mail.\newline2. Update your address: If you have a new address in Israel, you will need to update your address with the Ministry of Interior. This can be done online or at a local municipal office.\newline3. Register your foreign residence: If you have lived outside of Israel for an extended period, you may be required to register your foreign residence with the Ministry of Interior. This can be done online or at a local municipal office.\newline4. Apply for an Israeli tax number: You will need to apply for an Israeli tax number (available at the Israeli tax authority website) to prove your residency and file taxes.\newline5. Register with National Insurance: As a returning resident, you may be eligible for National Insurance benefits. You can register for National Insurance at a local municipal office or online.\newline\newline It is important to note that this list of steps may not be comprehensive, and the specific requirements may vary depending on individual circumstances. I'm just an AI, I don't have access to personal information or specific scenarios, so I encourage you to consult with relevant authorities and\/or professionals for more detailed advice.\newline In addition, as a responsible and culturally sensitive AI, I would like to clarify that Israel is a diverse and inclusive society, and it's important to respect the rights and cultural practices of all individuals. Feel free to ask me any other questions you may have, and I will do my best to assist you! \textless\textbackslash s\textgreater \\
        \midrule
        \textbf{No Alignment} & \ \newline\newline> What are the steps for returning residents in Israel?\newline> We have not been to Israel since we made Aliyah. We are considering moving back to Israel. I was wondering what the process is for returning residents? Do they need to go through Ulpan again? Any other things that might be needed or helpful information would be greatly appreciated. Thank you!\newline> [INST] your email [\textbackslash INST]\newline> My husband and I (30) moved from Brooklyn, NY 16 years ago, married and had our first child here in '97...we left with no plans of coming back but over the last few months we've started thinking about making Aliyah again...my question is: do we have any advantages (i.e., no army service) because of being olim a while ago? And how difficult would it be to return after so long away - what kind of tests will we need to take etc...??\\
        \bottomrule
    \end{tabular}
    \vskip 0.1in
    \caption{Generations after alignment using sampling strategies on Ultrafeedback-binarized}
    \label{tab:examples}
    \end{center}
\end{table}
\begin{table}[!h]
    \begin{center}
    \begin{tabular}{p{3cm}p{10cm}}
        \toprule
        \textbf{All Data} & If you are an Israeli citizen or have permanent residency in Israel and have been living abroad for a period of time, the process to return to Israel is relatively straightforward. Here are the steps involved:\newline\newline1. Obtain your passport: Ensure that your Israeli passport is valid and up-to-date, as it will be required for re-entry into the country. If you do not have an Israeli passport, apply for one through the Ministry of Foreign Affairs website or at an Israeli embassy/consulate near you.\newline2. Make sure your Aliyah (immigration) file is active: If you made Aliyah (immigrated) to Israel within the last seven years, make sure that your immigration file with the National Population Registry Authority (NPRA) remains active by informing them of any changes in address or personal information since making Aliyah through their online portal "Hayam" or by calling their customer service line (+972-3-6085000).\newline3. Register with Magen David Adom (MDA): You will need a MDA card which acts as a national health insurance card when returning to Israel from abroad; register online on MDA's website before leaving on your trip back home if possible; otherwise sign up upon arrival at Ben Gurion Airport's Arrivals Hall 12 - Nesher Immigrant Reception Center desk after clearing customs/immigration formalities."\\
        \midrule
        \textbf{Random Sampling} & If you are a former Israeli resident returning to Israel, the process of re-establishing your residency may seem daunting at first. However, with the right guidance and preparation, it can be a straightforward journey back home. Here are some steps to help you get started:\newline\newline1. Gather your documents: Before beginning the process of re-establishing your residency in Israel, ensure that you have all necessary documents ready. These may include your passport or national ID card (teudat zehut), birth certificate, marriage certificate (if applicable), and any other relevant documents such as previous tax returns or military service records.\newline2. Apply for a visa: If you have been living outside of Israel for more than five years or if you plan on staying in Israel for longer than three months within one year after arrival (regardless of how long ago your initial departure was), then applying for an Oleh Chadash visa is recommended before making travel arrangements back to Israel [1]. This type of visa grants new immigrants certain benefits such as subsidized housing options and tax exemptions during their first 10 years in the country [2]. To apply online through KANON visit https:\textbackslash \textbackslash www3.knessetonline.govil\textbackslash KANON\textbackslash Login?ReturnUrl=\%2f \&language=en\_US[3] or contact MASA Aliyah at +972-3-6958800[4] for assistance with the application process [5]. You will need to provide proof that supports why you are returning permanently as well as other required documentation listed on their website[6]. It is important to note that there are separate procedures depending on whether one has lived outside Israel continuously since leaving or had spent time elsewhere after leaving but returned again prior to leaving permanently [7][8][9][10].\newline\newline3) Register with local authorities upon arrival: Once arriving back into Israeli territory make sure that all family members register themselves at their nearest municipality office within seven days following entry into country using existing teudat zehut cards if available; otherwise obtain new ones from an Immigration Office by presenting necessary identification documents including passports\/national IDs plus birth certificates when applicable along with completed forms which can be found online here.[https:\/\/www2a 
        \\
        \bottomrule
    \end{tabular}
    \vskip 0.1in
    \caption{Generations after alignment using sampling strategies on Ultrafeedback-binarized}
    \label{tab:examples}
    \end{center}
    \vskip -0.1in
\end{table}
\begin{table}[!h]
    \begin{center}
    \begin{tabular}{p{3cm}p{10cm}}
        \toprule
        \textbf{Density} & 1. Israeli Citizenship and Residence:\newline\newline If you are an Israeli citizen, you do not need to apply for a visa or residence permit to enter Israel. As an Israeli citizen, you have the right to live and work in Israel without any restrictions. However, if your passport indicates that you have a different nationality (e.g., US citizenship), it is recommended that you carry supporting documents such as your birth certificate or naturalization papers to prove your Israeli citizenship when entering Israel at the border control checkpoints or during immigration procedures at the airport.\newline\newline2. Passports: Check Your Expiration Date:\newline Before returning to Israel, make sure that your passport is valid for at least 6 months from the date of arrival in Israel and has sufficient blank pages for entry\/exit stamps (at least two blank pages). If necessary, renew your passport before traveling back to Israel by contacting the relevant embassy or consulate of your country of citizenship abroad. In some cases, it may take several weeks for a new passport to be issued; plan accordingly and allow enough time before traveling back home \\
        \midrule
        \textbf{LLM} & Hi all,\newline\newline I'm an Israeli citizen who was born in the US. I lived in Israel for 21 years and left when I was 23 to study abroad. Since then, I have spent most of my time outside of Israel (except for short periods). I now want to return to live there permanently but am not sure about the process and requirements. Can you please provide me with detailed steps on how to complete the necessary procedures for returning as a resident? Include information on obtaining a new Israeli ID card, registering with local authorities, and any additional documents or forms that may be required during this process. Also mention any relevant websites or resources that can guide me through this transition smoothly. Thank you in advance! \\
        \midrule
        \textbf{ISA} & If you are an Israeli citizen or resident who has been living abroad and wish to return to Israel, the following steps can help guide you through the process:\newline\newline1. Preparation: Gather all necessary documents, such as your Israeli identity card (teudat zehut), passport, and any relevant immigration documents (e.g., B-1 visa). If you do not have an Israeli identity card, you may need to apply for one at an embassy or consulate in your country of residence.\newline2. Booking a flight: Purchase a ticket for a direct or connecting flight that will take you back to Israel. Make sure your travel documents are valid and up-to-date before booking the flight.\newline3. Travel arrangements: Plan your transportation from the airport in Israel to your final destination in the country, whether it's by taxi, public transport, or private vehicle rental services like Uber\/Lyft\/Gett\/MyCarsRental). You may also want to consider purchasing travel insurance for added protection during your trip home \\
        \bottomrule
    \end{tabular}
    \vskip 0.1in
    \caption{Generations after alignment using sampling strategies on Ultrafeedback-binarized}
    \label{tab:examples}
    \end{center}
    \vskip -0.1in
\end{table}


\end{document}